\documentclass[10pt,twocolumn,letterpaper]{article}

\usepackage{cvpr}
\usepackage{times}
\usepackage{epsfig}
\usepackage{graphicx}
\usepackage{amsmath}
\usepackage{amssymb}
\usepackage{dsfont}
\usepackage{subfig}
\usepackage{cite}
\usepackage{authblk}
\usepackage{colortbl}
\usepackage[table]{xcolor}

\usepackage[pagebackref=true,breaklinks=true,letterpaper=true,colorlinks,bookmarks=false]{hyperref}

\definecolor{TableRed}{HTML}{800000}
\newcommand{\TextRed}[1]{\textcolor{TableRed}{#1}}

\cvprfinalcopy % *** Uncomment this line for the final submission

\def\cvprPaperID{****} % *** Enter the CVPR Paper ID here

\ifcvprfinal\fi
\begin{document}

%%%%%%%%% TITLE
\title{Deep Texture Manifold for Ground Terrain Recognition}

\author[1]{
Jia Xue
}
\author[1,2]{
Hang Zhang
}
\author[1]{
Kristin Dana
}
%\affil[1]{Department of Electrical and Computer Engineering, Rutgers University, New Brunswick, USA}
%\affil[]{Amazon AI}
\affil[ ]{$^1$Department of Electrical and Computer Engineering, Rutgers University, $^2$Amazon AI}
\affil[ ]{ {\tt\small \{jia.xue, zhang.hang\}@rutgers.edu,  kdana@ece.rutgers.edu }}
\renewcommand\Authsep{  } 
\renewcommand\Authands{  }

\maketitle
%\thispagestyle{empty}

%%%%%%%%% ABSTRACT
\begin{abstract}
   We present a texture network called Deep Encoding Pooling Network (DEP)   for the task of ground terrain recognition. Recognition of ground terrain is an important task in establishing robot or vehicular control parameters, as well as for localization within an outdoor environment.  The architecture of DEP integrates orderless texture details and local spatial information and the performance of DEP surpasses state-of-the-art methods for this task. The GTOS database (comprised of over 30,000 images of 40 classes of  ground terrain in outdoor scenes)  enables  supervised recognition. 
   %However, instead of using the database in a standard way  (splitting the image set for testing and evaluation), we assume no access to the original camera sensor, nor geometric/photometric conditions of the GTOS database. 
For evaluation under realistic conditions, we use test images that are not from 
the existing GTOS dataset, but are instead 
from hand-held mobile phone videos of similar terrain. 
This new evaluation dataset, GTOS-mobile, consists of 81 videos of  31 classes of ground terrain such as grass, gravel, asphalt and sand.
The resultant network shows
excellent performance not only for GTOS-mobile, but also for more general databases (MINC and DTD).
% A texture manifold of ground terrain images is important for recognition evaluation, since there are often  ambiguous class boundaries.
Leveraging the discriminant features learned from this network, we build a new texture manifold called DEP-manifold. 
We learn a parametric distribution in feature space  
%The position of an image within this manifold is learned directly 
in a fully supervised manner, which gives the distance relationship among classes and provides a means to implicitly represent ambiguous class boundaries. The source code and database are publicly available\footnote{\url{http://ece.rutgers.edu/vision/}}.
\end{abstract}

%%%%%%%%% BODY TEXT
\section{Introduction}

%  In the field of material recognition and computational photography, one goal is to develop a ground terrain recognition system which can recognize ground terrain surfaces and find the relationship between newly captured material images and the corresponding images in the material dataset.
   
\begin{figure}[t]
\centering
\subfloat
{
\includegraphics[width=.3\linewidth]{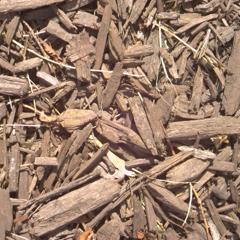}
}
\subfloat
{
\includegraphics[width=.3\linewidth]{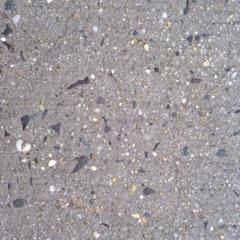}
}
\subfloat
{
\includegraphics[width=.3\linewidth]{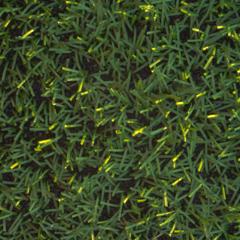}
}
\setcounter{subfigure}{0}
\subfloat
{
\includegraphics[width=.3\linewidth]{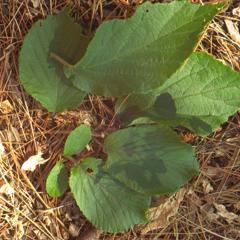}
}
\subfloat
{
\includegraphics[width=.3\linewidth]{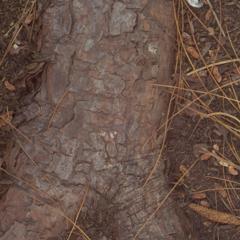}
}
\subfloat
{
\includegraphics[width=.3\linewidth]{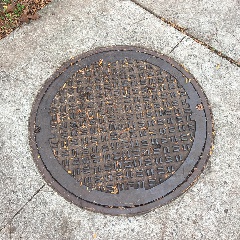}
}
\caption{Homogeneous textures  (upper row) compared to more common real-world instances with local spatial structure that provides an important cue for  recognition (lower row).
% such as leaves, roots and metal cover. 
}
\label{fig:compare_img_patch}
\vspace{-0.1in}
\end{figure}

\begin{figure*}[t]
\centering
\includegraphics[width= \linewidth]{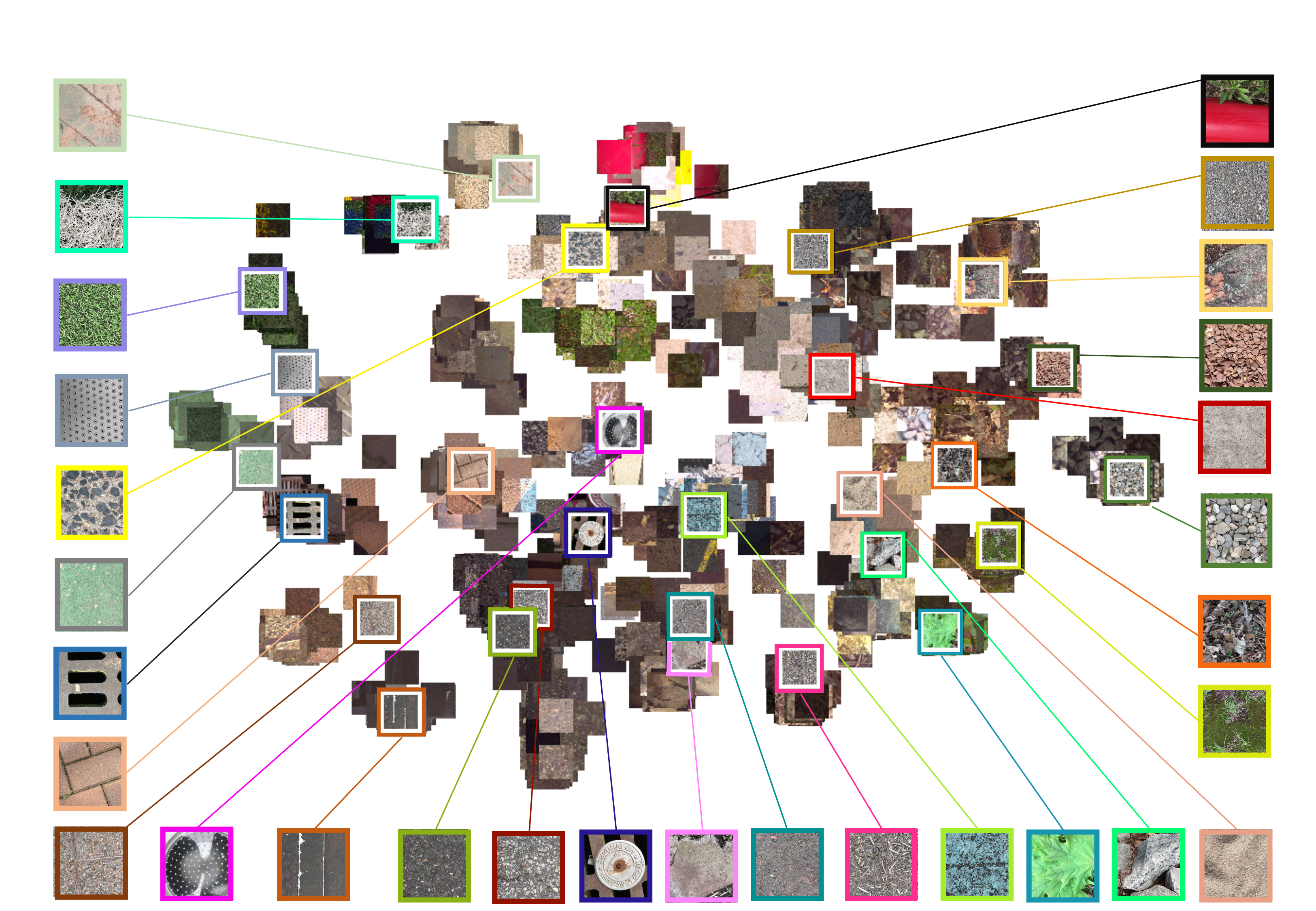}
\caption{The result of texture manifold by DEP-manifold. Images with color frames are images in test set. The material classes are (from upper left to counter-clockwise): plastic cover, painted turf, turf, steel, stone-cement, painted cover, metal cover, brick, stone-brick, glass, sandpaper, asphalt, stone-asphalt, aluminum, paper, soil, mulch, painted asphalt, leaves, limestone, sand, moss, dry leaves, pebbles, cement, shale, roots, gravel and plastic. Not all classes are shown here for space limitations.}
\label{fig:manifold}
\vspace{-0.1in}
\end{figure*}

   Ground terrain recognition is  an  important area of research in computer vision for potential applications in autonomous driving and robot navigation. Recognition with CNNs  have achieved  success in object recognition and the CNN  architecture balances preservation of relative spatial information (with convolutional layers) and aggregation of spatial information (pooling layers). This structure is designed for object recognition, scene understanding, face recognition, and applications where spatial order is critical for classification. However,  texture recognition uses an orderless  component  to provide invariance to  spatial layout \cite{zhang2016deep,lin2013network,cimpoi2015deep}.
   
   In classic approaches for texture modeling, images are filtered with a set of handcrafted filter banks 
   %such as SIFT \cite{lowe2004distinctive}
followed by grouping
 the outputs into texton histograms  \cite{Malik2001, Cula01,leung2001representing,Varma2005}, or bag-of-words \cite{Csurka04,Lazebnik06}.
 %is employed to detect interesting edges and points as material appearance distributions. The detected material appearance distributions are encoded by dictionary learning algorithms \cite{perronnin2010improving,jegou2010aggregating} for recognition. 
 Later, Cimpoi \etal \cite{cimpoi2015deep} introduce FV-CNN that replace the handcrafted filter banks with  pre-trained convolutional layers for the feature extractor, and achieve state-of-the-art results. Recently, Zhang \etal  \cite{zhang2016deep} introduce Deep Texture Encoding Network (Deep-TEN) that ports the dictionary learning and feature pooling approaches into the CNN pipeline for an end-to-end material/texture recognition network.
Recognition algorithms that focus on texture details work well for images containing only a single material. But for ``images in the wild", homogeneous surfaces rarely fill the entire field-of-view, and many materials exhibit regular structure. 

For texture recognition,  since surfaces are not completely orderless, {\it local spatial order}  is an important cue for recognition as illustrated in Figure~\ref{fig:compare_img_patch}.
Just as semantic segmentation balances local details and global scene context for pixelwise recognition \cite{Zheng_2015_ICCV,Mottaghi_2014_CVPR,Lin17exploring,schwartz2016material,bell2015material,Shelhamer17}, 
we  design a network to balance both an orderless component and ordered spatial information.
% The spatial information is equally as important as texture details for real world material recognition. This is similar to humans identify materials, we first look at the shape and appearance of an object, and if we cannot distinguish the material class, we will take a closer look at the texture details. 

\begin{figure*}
\centering
\includegraphics[width= \linewidth]{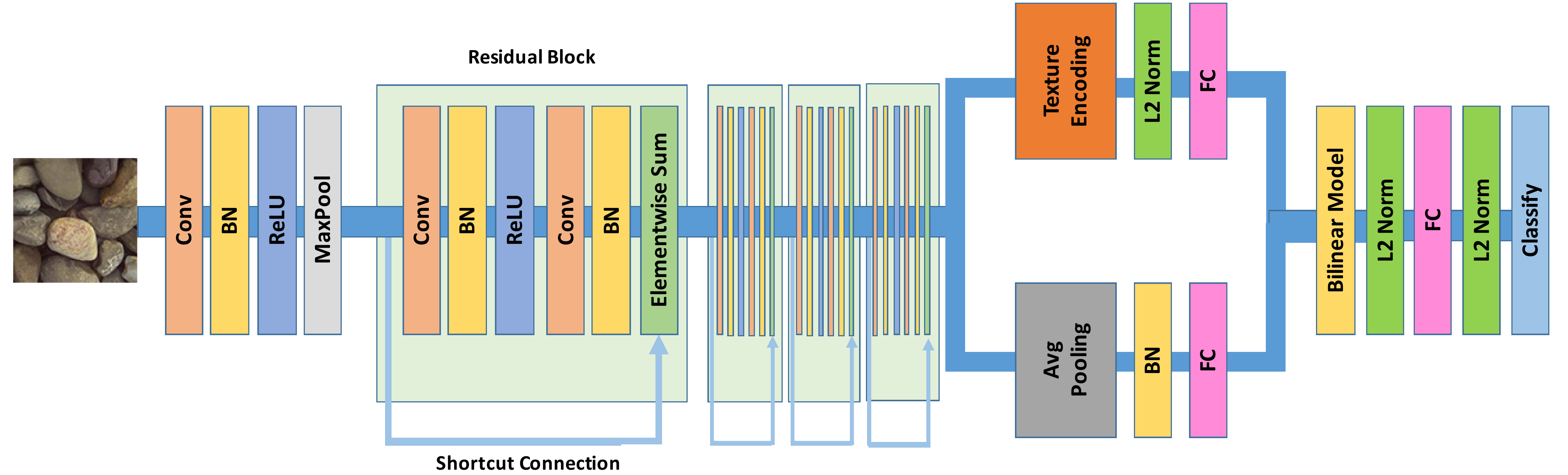}
\caption{A Deep Encoding Pooling Network (DEP) for material recognition. Outputs from convolutional layers are fed into  the encoding layer and global average pooling layer jointly and their outputs are processed with bilinear model.}
\label{fig:algorithm}
% \vspace{-0.2in}
\end{figure*}

As illustrated in  Figure ~\ref{fig:algorithm}, we introduce a Deep Encoding Pooling Network (DEP) that leverages an orderless representation and local spatial information for recognition. 
   Outputs from convolutional layers are fed into two feature representation layers jointly; the  encoding layer \cite{zhang2016deep} and the global average pooling layer.
   The encoding layer is employed to capture texture appearance details and the global average pooling layer accumulates spatial information.
   Features from the encoding layer and the global average pooling layer are processed with bilinear models \cite{tenenbaum1997separating}.  We apply DEP to the problem of ground terrain recognition using an extended GTOS dataset \cite{xue2016differential}.
 The resultant network shows excellent performance not only for GTOS, but also for more general databases (MINC \cite{bell2015material} and DTD \cite{cimpoi2014describing}).

For ground terrain recognition, many class boundaries are ambiguous. For example, ``asphalt'' class is similar to ``stone-asphalt'' which is an aggregate mix of stone and asphalt. The class ``leaves'' is similar to ``grass'' because most of the example images for ``leaves'' in the GTOS database have grass in the background. Similarly, the grass images contain a few leaves.  
Therefore, it is of interest to find not only the class label but also the closest classes, or equivalently, the position in the manifold.  
We  introduce a new texture manifold method, DEP-manifold, to find the relationship between newly captured images and images in dataset.

The t-Distributed Stochastic Neighbor Embedding (t-SNE)
\cite{maaten2008visualizing}  provides a  2D embedding and   Barnes-Hut t-SNE \cite{van2014accelerating} accelerates the original t-SNE from $\mathcal{O}({n}^{2})$ to $\mathcal{O}(n\log{}n)$. 
Both t-SNE  and and Barnes-Hut t-SNE are non-parametric embedding algorithms,  so there is no natural way to perform out-of-sample extension. Parametric t-SNE \cite{van2009learning} and supervised t-SNE \cite{min2010deep,min2017exemplar} introduce deep neural networks into data embedding and realize non-linear parametric embedding. Inspired by this work, we introduce a method for texture manifolds that treats the embedded distribution from non-parametric embedding algorithms as an output, and use a deep neural network to predict the manifold coordinates of a texture image directly. This texture manifold uses the features of the DEP network and is referred to as DEP-manifold.
%With this method, the  data distribution is computed by the non-parametric embedding, and the deep neural networks are employed to fit the computed distribution. 

    The training set is a ground terrain database (GTOS) \cite{xue2016differential} with 31 classes of ground terrain images (over 30,000 images in the dataset). Instead of using images from the GTOS dataset for testing,  we collect GTOS-mobile, 81 ground terrains videos of similar terrain classes captured with a  hand-held mobile phone and with arbitrary lighting/viewpoint. Our motivation is as follows: The training set (GTOS) is obtained in a comprehensive manner (known distance and viewpoints, high-res caliabrated camera) and is used to obtain knowledge of the scene. The test set is obtained under very different and more realistic conditions (a mobile imaging device, handheld video, uncalibrated capture).
    %The application envisioned is a robot or automated vehicle traversing an area from which training images were once obtained but with a different sensor, weather, viewpoint and illumination. 
     Training with GTOS and testing with GTOS-mobile enables evaluation of knowledge transfer of the network. 

\section{Related Work}

   Tenenbaum and Freeman \cite{tenenbaum1997separating} introduce bilinear models to process two independent factors that underly a set of observations. %Recently, 
   Lin \etal \cite{lin2015bilinear} introduce the Bilinear CNN models that use outer product of feature maps from convolutional layers of two CNNs and reach state-of-the-art for fine grained visual recognition. However, this method has two drawbacks. First, bilinear models for feature maps from convolutional layers require that pairs of features maps have compatible feature dimensions, i.e.\ the same height and width. % and channel numbers. %% channels can be different  %%
   The second drawback is computational complexity; this method computes the outer product at each location of the feature maps. To utilize the advantage of bilinear models and overcome these drawbacks, we  employ bilinear models for outputs from fully connected layers. Then, outputs from fully connected layers can be treated as vectors, and there is no dimensionality restriction for the outer product of two vectors.

Material recognition is a fundamental problem in computer vision, the analysis of material recognition has varied from small sets  collected in lab settings such as  KTH-TIPS \cite{caputo2005class} and CuRET \cite{dana1999reflectance}, to large image sets collected in the wild \cite{wang20164d,bell2015material,cimpoi2014describing,xue2016differential}. The size of material datasets have also increased from roughly 100 images in each class \cite{wang20164d,cimpoi2014describing} to over 1000 images in each class \cite{xue2016differential,bell2015material}.
 % Many datasets \cite{dana1999reflectance,caputo2005class, wang20164d} are captured with one camera and are settled to  camera parameters: white balance, exposure time, lens distortion, etc. We propose the challenge that whether computer can recognize materials obtained with a different sensor, weather, viewpoint and illumination based on the knowledge learned from one camera. Recently Xue \etal  introduce the
 The Ground Terrain in Outdoor Scenes (GTOS) dataset has been used with angular differential imaging \cite{xue2016differential}  for  material recognition based on angular gradients. 
 For our work,  single images are used for recognition without variation in viewing direction, so reflectance gradients are not considered. 
 %instead of dividing this dataset into test and training images, we take uncalibrated images with a handheld mobile phone to replicate a realistic recognition scenario. We train our network with the comprehensive high-res images of the GTOS dataset but then evaluate with mobile video images of similar terrain (GTOS-mobile). 

 For many recognition problems, deep learning has achieved great success, such as face recognition  \cite{zhang2017generative,meng2017identity,cai2017island,peng2016recurrent}, action recognition \cite{zhu2016depth2action,zhu2017hidden} and disease diagnosis \cite{zhang2017mdnet}.
 The success of deep learning has also transferred to material recognition. We leverage a recent texture encoding layer \cite{zhang2016deep} 
 that ports dictionary learning and residual encoding into CNNs. We use this texture encoding layer as a component in our network to capture orderless texture details.

%  {\bf Data Visualization} Hinton and Roweis \cite{hinton2003stochastic} introduce Stochastic Neighbor Embedding (SNE) that projects the high dimensional Euclidean distance between data points into conditional probabilities and show similarity between high dimensional data points with low dimensional representation. Later on Matten and Hinton \cite{maaten2008visualizing} introduce t-Distributed Stochastic Neighbor Embedding (t-SNE) that uses Student-t distribution to compute similarities between two points and visualize high dimensional data distribution with 2D or 3D images. Later Matten \cite{van2014accelerating} introduce Barnes-Hut t-SNE to optimize t-SNE computation complexity. However, for both methods, there are no parameters to interpret the relationship between high dimensional data and low dimensional manifold, so there is no natural way to perform out of sample extension. To solve the problem, Matten \cite{van2009learning} employ deep neural networks for data embedding and propose parametric t-SNE, which is a three-stage training procedure. Supervised t-SNE \cite{min2010deep,min2017exemplar} models the similarities of data belonging to the same class in  the  embedding, they treat the conditional probability as ``ground truth'' to train the deep neural network.  

\section{Deep Encoding Pooling Network}

    %We are interested in exploring the problem of transferring  knowledge learned from images captured by one camera to images captured by another camera. A new dataset is created based on the GTOS database \cite{xue2016differential}, which contains over 30000 ground terrain images captured by one camera.
   %Additionally we collect 81 material videos from similar material areas with an iphone SE.
   %To design a material collection camera that able to scan ground terrain and store terrain information for future retrieval and recognition.
   %We introduce binary texture code that compress texture images into binary codes, the compressed binary code times the trained classification layer with get the image information.
  %In this way, we combine the compact efficiency of binary codes with robust performance of deep learning.
  
 % In this section we first introduce a {\it Deep Texture Encoding and Bilinear Module} for texture recognition, then introduce the {\it Deep Encoding Pooling Network (DEP)}, and describe a specific instantiation of DEP. 
  
  \paragraph{Encoding Layer} The texture encoding layer~\cite{zhang2016deep} integrates the entire dictionary learning and visual encoding pipeline into a single CNN layer, which provides an orderless representation for texture modeling. 
The encoding layer acts as a global feature pooling on top of convolutional layers. 
Here we briefly describe prior work for completeness. %is an orderless feature pooling layer that works as the dictionary learning and feature pooling in classic material recognition approaches.  %is integrated on top of convolution layers and represents  outputs from convolution layer with learnable codewords. 
  Let $X = \left \{ {x}_{1},...{x}_{m} \right \}$ be M visual descriptors, $C = \left \{ {c}_{1},...{c}_{n} \right \}$ is the code book with N learned codewords. The residual vector ${r}_{ij}$ is calculated by ${r}_{ij} = {x}_{i} - {c}_{j}$, where $i = 1 ...m$ and $j = 1 ...n$.
  The residual encoding for codeword ${c}_{j}$ can be represented as 
  \begin{equation}
  {e}_{j} = \sum_{i=1}^{N} {w}_{ij}{r}_{ij} ,
  \end{equation}
  where ${w}_{ij}$ is the assigning weight for residual vector ${r}_{ij}$ and is given by
  \begin{equation}
  {w}_{ij} = \frac{\exp(-{s}_{j}{\lVert{r}_{ij}\rVert}^{2})}{\sum_{k=1}^{m} \exp({-{s}_{k}{\lVert{r}_{ik}\rVert}^{2}})} ,
  \end{equation}
  $ {s}_{1},...{s}_{m}$ are learnable smoothing factors. With the texture encoding layer, the visual descriptors X are pooled into a set of N residual encoding vectors $E = \left \{ {e}_{1},...{e}_{n} \right \}$. 
  % "we can find" is not very formal
  Similar to classic encoders, the encoding layer can capture more texture details by increasing the number of learnable codewords.
  
  \paragraph{Bilinear Models}  
  Bilinear models are two-factor models such that their outputs are linear in one factor if the other factor is constant~\cite{freeman1997learning}.
  The factors in bilinear models balance the contributions of the two components. 
  Let ${a}^{t}$ and ${b}^{s}$ represent the material texture information and spatial information with vectors of parameters and with dimensionality $I$ and $J$. The bilinear function $Y^{ts}$ is given by 
  \begin{equation}
  Y^{ts} = \sum_{i=1}^{I}\sum_{j=1}^{J} {w}_{ij}{a}_{i}^{t}{b}_{j}^{s} ,
  \end{equation}
  where ${w}_{ij}$ is a learnable weight to balance the interaction between material texture and spatial information.  The outer product representation captures a pairwise correlation between the material texture encodings and spatial observation structures.
  
  \paragraph{Deep Encoding Pooling (DEP) Network} Our Deep Encoding Pooling Network (DEP) is shown in Figure~\ref{fig:algorithm}. As in prior transfer learning algorithms \cite{lin2015bilinear,zhang2016deep}, we employ  convolutional layers with non-linear layers from ImageNet \cite{deng2009imagenet} pre-trained CNNs as feature extractors.  Outputs from convolutional layers  are fed into the texture encoding layer and the global average pooling layer jointly. Outputs from the texture encoding layer preserve  texture details, while outputs from the global average pooling layer preserve local spatial information. The dimension of outputs from the texture encoding layer is determined by the codewords N and the feature maps channel C (N$\times$C). The  dimension of outputs from the global average pooling layer is determined by the  feature maps channel C. For computational efficiency and to robustly combine feature maps with bilinear models, we reduce feature maps dimension with fully connected layers for both branches. Feature maps from the texture encoding layer and the global average pooling layer are processed with a bilinear model and followed by a fully connected layer and a classification layer with non-linearities for classification. Table ~\ref{table:structure} is an instantiation of DEP based on 18-layer ResNet \cite{he2016deep}. We set 8 codewords for the texture encoding layer. The size of input images are $224\times224$. Outputs from CNNs are fed into the texture encoding layer and the global average pooling layer jointly. The dimension of outputs from the texture encoding layer is $8\times512=4096$ and the dimension of outputs from global average pooling layer is 512.
  We reduce the dimension of feature maps from the deep encoding layer and the global average pooling layer to 64 via fully connected layers. The dimension of outputs from bilinear model is $64\times64=4096$.
  Following prior works \cite{zhang2016deep,perronnin2010improving}, resulting vectors from the texture encoding layer and bilinear model are normalized with L2 normalization.
 
  The texture encoding layer and bilinear models are both differentiable.
  The overall architecture is a directed acyclic graph and all the parameters can be trained by back propagation.
  Therefore, the Deep Encoding Pooling Network is trained end-to-end using stochastic gradient descent with back-propagation.

{\setlength{\extrarowheight}{3pt}%
\begin{table}
\centering
\resizebox{\linewidth}{!}{%
\def\arraystretch{1.3}
\begin{tabular}{|l|c|c|}
\hline
layer name & output size & encoding-pooling \\ \hline
conv1      & 112$\times$112$\times$64  & 7$\times$7, stride 2    \\ 
\hline
\rule{0pt}{20pt}conv2\_x       & 56$\times$56$\times$64   &  $\begin{bmatrix} 3 \times 3, 64\\ 3 \times 3, 64 \end{bmatrix} \times 2$    \\ [1em]
\hline
\rule{0pt}{20pt}conv3\_x      & 28$\times$28$\times$128   &  $\begin{bmatrix} 3 \times 3, 128\\ 3 \times 3, 128 \end{bmatrix} \times 2$    \\ [1em]
\hline 
\rule{0pt}{20pt}conv4\_x      & 14$\times$14$\times$256   &  $\begin{bmatrix} 3 \times 3, 256\\ 3 \times 3, 256 \end{bmatrix} \times 2$    \\ [1em]
\hline 
\rule{0pt}{20pt}conv5\_x      & 7$\times$7$\times$512   &  $\begin{bmatrix} 3 \times 3, 512\\ 3 \times 3, 512 \end{bmatrix} \times 2$    \\ [1em]
\hline
encoding / pooling & 8 x 512 /  512          &   8 codewords / ave pool   \\ 
\hline 
fc1\_1 / fc1\_2 & 64 / 64  & 4096$\times$64 / 512$\times$64   \\ 
\hline 
bilinear mapping & 4096  &  -   \\ 
\hline 
fc2 & 128 & 4096$\times$128   \\ 
\hline
classification & n classes & 128$\times$n\\
\hline

\end{tabular}
}
\caption{The architecture of Deep Encoding Pooling Network based on 18-layer ResNet \cite{he2016deep}. The input image size is $224 \times 224$.}
%\vspace{-0.2in}
\label{table:structure}
\end{table}

\section{Recognition Experiments}
%\subsection{Methods} 
%\label{Methods}

\begin{figure*}[t]
\centering
\subfloat
{
\includegraphics[width=.47\linewidth]{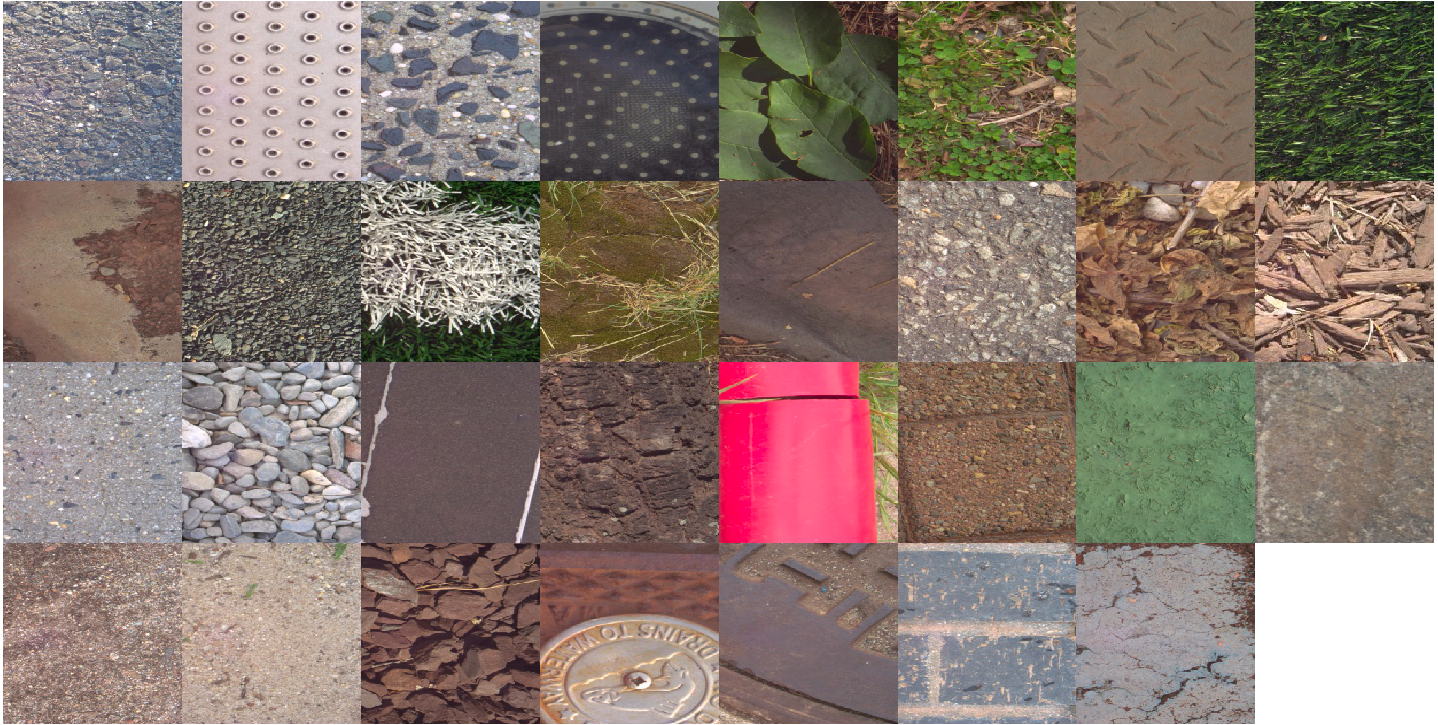}
}
\hspace{3mm}
\subfloat
{
\includegraphics[width=.476\linewidth]{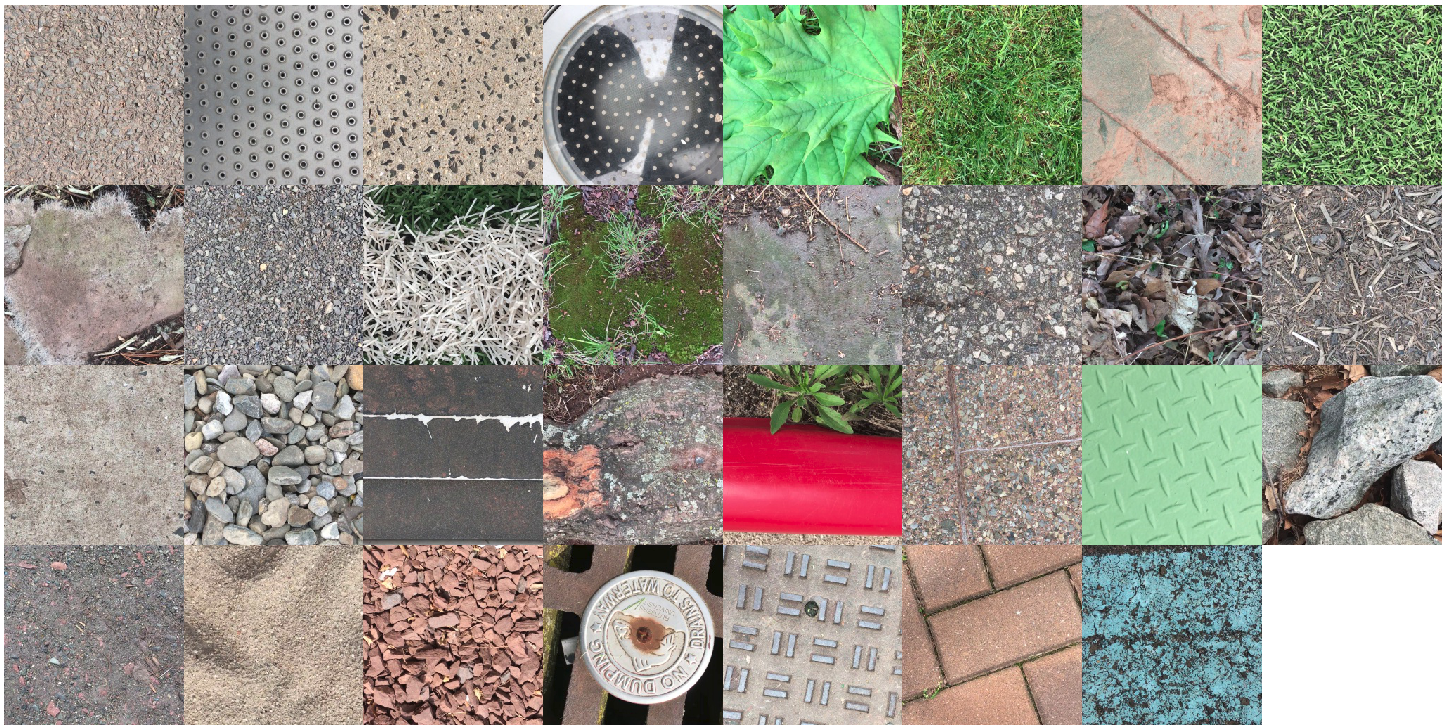}
}
\caption{Comparison of images from the GTOS  dataset (left)  and GTOS-mobile (right) video frames. The training set is the ground terrain database (GTOS)
with 31 classes of ground terrain images (over 30,000 images
in the dataset).  GTOS is collected with  calibrated viewpoints. GTOS-mobile,  consists of 81 videos
of similar terrain classes captured with a handheld
mobile phone and with arbitrary lighting/viewpoint. A total of  6066 frames are extracted from the videos with a temporal sampling of approximately 1/10th seconds. The figure shows individual frames of 31 ground terrain classes.}
\label{fig:database}
%\vspace{-0.2in}
\end{figure*}

  We compare the DEP network with the following three methods based on ImageNet \cite{russakovsky2015imagenet} pre-trained 18-layer ResNet \cite{he2016deep}: (1) CNN with ResNet, (2) CNN with Deep-Ten and(3) CNN with bilinear models. 
  All three methods support end-to-end training. For equal comparison, we use an identical training and evaluation procedure  for each experiment.
  \paragraph{CNN with global average pooling (ResNet)}
  %Since a 7$\times$7 global average pooling layer is already in the 18-layer pre-trained ResNet. 
  We follow the standard procedure to fine-tune pre-trained ResNet, by replacing the last 1000-way fully connected layer with the output dimension of 31. 
  The global average pooling works as feature pooling that encodes the 7$\times$7$\times$512 dimensional features from the 18-layer pre-trained ResNet into a 512 dimensional vector. 
  %In the 18-layer pre-trained ResNet, an 1000-way classification layer follows the global average pooling layer for classification. We replace the 1000-way classification layer with a new classification layer, the output dimension of new classification layer is 31.

  \paragraph{CNN with texture encoding (Deep-TEN)}
  The Deep Texture Encoding Network (Deep-TEN) \cite{zhang2016deep} embeds the texture encoding layer on top of the 50-layer pre-trained ResNet \cite{he2016deep}. To make an equal comparison, we replace the 50-layer ResNet with 18-layer ResNet.
  Same as \cite{zhang2016deep}, we reduce the number of CNN streams outputs channels from 512 to 128 with a 1$\times$1 convolutional layer. We replace the global average pooling layer in the 18-layer ResNet with texture encoding layer, set the number of codewords to 32 for experiments. Outputs from the texture encoding layer are normalized with L2 normalization. A fully connected layer with soft max loss follows the texture encoding layer for classification.

  \paragraph{CNN with bilinear models (Bilinear-CNN)}
Bilinear-CNN \cite{lin2015bilinear}  employs bilinear models with feature maps from convolutional layers. Outputs from convolutional layers of two CNN streams are multiplied using outer product at each location and pooled for recognition. To make an equal comparison, we employ the 18-layer pre-trained ResNet as CNN streams for feature extractor. Feature maps from the last convolutional layer are pooled with bilinear models. The dimension of feature maps for bilinear models is 7$\times$7$\times$512 and the pooled bilinear feature is of size 512$\times$512. The pooled bilinear feature is fed into classification layer for classification.

\begin{table*}[t]
\centering
\begin{tabular}{|l|l|l|l|l|}
\hline
 		& ResNet \cite{he2016deep} & Bilinear CNN\cite{lin2015bilinear} &	Deep-TEN\cite{zhang2016deep} & DEP ({\small\TextRed{ours}}) \\ \hline
 Single scale   & 70.82  & 72.03  &74.22 &\textbf{76.07} \\ \hline
 Multi scale & 73.16 & 75.43 & 76.12 & \textbf{82.18} \\ \hline
\end{tabular}
\caption{Comparison our Deep Encoding Pooling Network (DEP) with ResNet (left) \cite{he2016deep}, Bilinear CNN (mid) \cite{lin2015bilinear} and Deep-TEN (right) \cite{zhang2016deep} on GTOS-mobile dataset with single scale and multi scale training. For ResNet, we replace the 1000-way classification layer with a new classification layer, the output dimension of new classification layer is 31.}
\label{table:gtos}
%\vspace{-0.1in}
\end{table*}

\subsection{Dataset and Evaluation}
\paragraph{Dataset}
  %To measure the ability to adapt texture knowledge learned from images collected by one camera to images collected by other cameras.
  Extending the GTOS database \cite{xue2016differential}, we  collect {\it GTOS-mobile} consisting of 81  videos obtained with a mobile phone (Iphone SE)  and extract 6066 frames as a test set. To simulate real world ground terrain collection, we walk through similar ground terrain regions in random order to collect the videos. Scale is changed arbitrarily by  moving far or  close and changes in viewing direction are obtained by  motions in a small arc. The resolution of the videos is 1920$\times$1080, and we resize the short edge to 256 while keeping the aspect ratio for experiments. As a result, the resolution of the resized images are 455$\times$256.
 Some materials in GTOS were not accessible due to weather, therefore we removed the following classes: dry grass, ice mud, mud-puddle, black ice and snow from the GTOS dataset.
 Additionally, we merged very similar classes of asphalt and metal. The original GTOS set is 40 classes, as 
 shown in Figure~\ref{fig:database}, there are 31 material classes in the modified dataset. The class names are (in the order of top-left to bottom-right): asphalt, steel, stone-cement, glass, leaves, grass, plastic cover,  turf, paper, gravel, painted turf, moss, cloth, stone-asphalt, dry leaves, mulch, cement, pebbles, sandpaper, roots, plastic, stone-brick, painted cover, limestone, soil, sand, shale, aluminum, metal cover, brick, painted asphalt.
  
  \paragraph{Multi-scale Training}
  Images in the GTOS dataset were captured from a fixed distance between the camera and ground terrain,
  however the distance between the camera and ground terrain can be arbitrary in real world applications.
  We infer that extracting different resolution patches with different aspect ratio from images in GTOS simulate observing materials at different distance and viewing angle will be helpful for recognition.
  So for image pre-processing, instead of directly resizing the full resolution images into 256$\times$256 as \cite{xue2016differential}, we resize the full resolution images into different scales, and extract 256$\times$256 center patches for experiment. Through empirical validation, we find that resizing the full resolution images into 256$\times$256, 384$\times$384 and 512$\times$512 works best. 
 
% add encoding experiment here
  %\paragraph{Binary Texture Code}
  %We explore the method that use binary hashing to store scanned ground terrain materials.
  %Binary representation can save storage memory.
  %The binary representation we employ is that we add an extra fully connected layer after bilinear combination to shrink the output dimension, the parameters of the fully connected layer is initialized with Xavier initialization \cite{glorot2010understanding}.
  %To generate a binary like output, a batch normalization is followed the added fully connected layer.
  %For the reason that after batch normalization, each minibatch is normalized to mean zero, standard derivation one. 
  %We convert feature maps into binary like output with the definition that 
  %\begin{equation}
  %f(x) = \begin{cases}
 %1 & \text{ if } x \geq 0 \\ 
 %0 & \text{ if } x < 0 
 % \end{cases}
  %\end{equation}
  %We add a classification layer after the binary like function with xavier initialization. 
  %we retrain the encoding layer fully connected layer and classification layer to fit the binary like function.
  %We experiment with output dimension 64 and 32 of the added fully connected layer after bilinear combination, the result is shown on \ref{tabel:gtos}.
  %With this adjustment, our architecture is still end-to-end training.

  \paragraph{Training procedure}
  We employ an identical data augmentation and training procedure  for experiments.
  For single scale training experiment, we resize the full resolution images into 384$\times$384 and extract 256$\times$256 center patches as training set. For multi scale training experiment, we resize the full resolution images into 256$\times$256, 384$\times$384 and 512$\times$512, and extract 256$\times$256 center patches as training set.
  For the training section data augmentation, following prior work \cite{huang2016densely, zhang2016deep}, we crop a random size (0.8 to 1.0) of the original size and a random aspect ratio (3/4 to 4/3) of the original aspect ratio, resize the cropped patches to 224$\times$224 for experiment.
  All images are pre-processed by subtracting a per color channel mean value and normalized to unit variance with a 50\% chance horizontal flip.
  The learning rate of newly added layers is 10 times of the pre-trained layers. The experiment starts with learning rate at 0.01, momentum 0.9, batch size 128; the learning rate decays by factor of 0.1 for every 10 epochs, and is finished after 30 epochs.

\begin{figure*}[t]
\centering
\subfloat
{
\includegraphics[width=.33\linewidth]{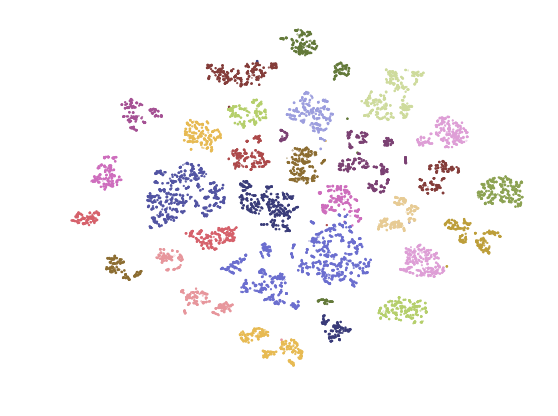}
}
\subfloat
{
\includegraphics[width=.33\linewidth]{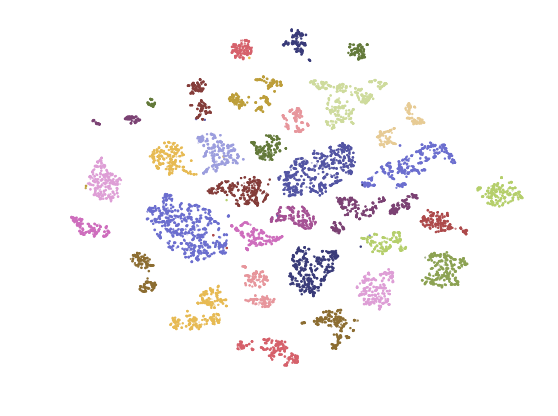}
}
\subfloat
{
\includegraphics[width=.33\linewidth]{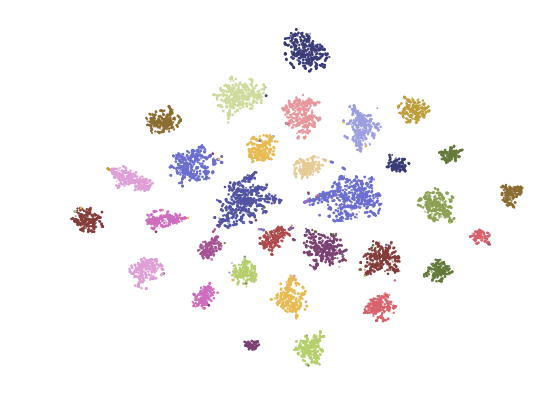}
}
\setcounter{subfigure}{0}
\subfloat[ResNet]
{
\includegraphics[width=.33\linewidth]{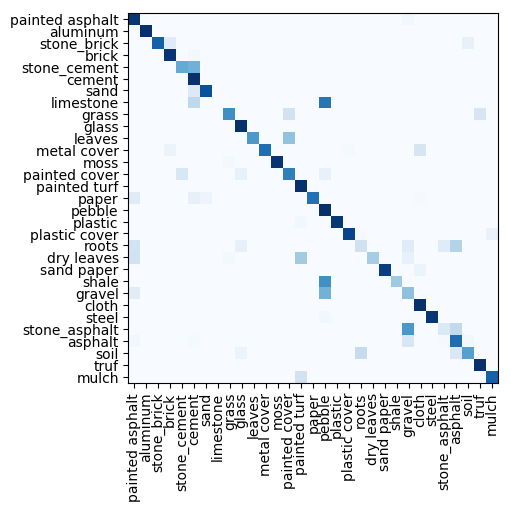}
}
% * <zhang.hang@rutgers.edu> 2017-11-15T17:31:03.679Z:
%
% ^.
\subfloat[Deep-TEN]
{
\includegraphics[width=.33\linewidth]{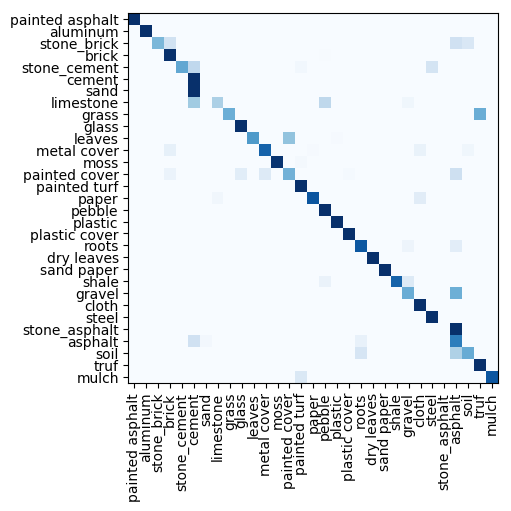}
}
\subfloat[DEP ({\small\TextRed{ours}})]
{
\includegraphics[width=.33\linewidth]{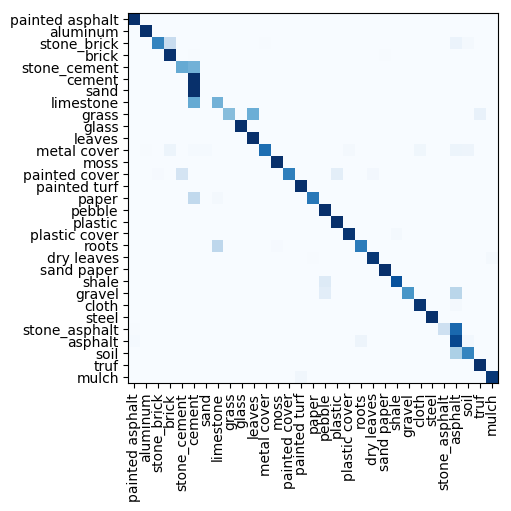}
}
\caption{The Barnes-Hut t-SNE \cite{van2014accelerating} and confusion matrix of three material recognition models: ResNet (left), Deep-TEN (mid) and DEP (right). For Barnes-Hut t-SNE,  we randomly choose 10000 images from training set and extract features before classification layers of three models for experiment. We see that DEP separates and clusters the classes better. Some classes are misclassified, however, they are typically recognized as a nearby class. (Dark blue represents higher values and light blue represents lower values in the confusion matrix.)}
\label{fig:tsne}
%\vspace{-0.1in}
\end{figure*}

\begin{table}[t]
\centering
\begin{tabular}{|c|c|c|}
\hline
Method & DTD\cite{cimpoi2014describing}    & Minc-2500\cite{bell2015material} \\ \hline
FV-CNN~\cite{cimpoi2015deep} & 72.3\% & 63.1\%    \\ \hline
Deep-TEN~\cite{zhang2016deep} & 69.6\% & 80.4\%    \\ \hline
DEP ({\small\TextRed{ours}})    & \textbf{73.2}\% & \textbf{82.0}\%    \\ \hline
\end{tabular}
\caption{Comparison with state-of-the-art algorithms on Describable Textures Dataset (DTD) and Materials in Context Database (MINC).}
\label{table:state-art}
%\vspace{-0.2in}
\end{table}

\subsection{Recognition Results}
  \paragraph{Evaluation on GTOS-mobile } Table \ref{table:gtos} is the classification accuracy of fine-tuning ResNet~\cite{he2016deep}, bilinear CNN~\cite{lin2015bilinear}, Deep-TEN~\cite{zhang2016deep} and the proposed DEP on the GTOS-mobile dataset. When comparing the performance of single-scale and multi-scale training, multi-scale training outperforms single-scale training for all approaches. It proves our inference that extracting different resolution patches with different aspect ratio from images in GTOS to simulate observing materials at different distance and viewing angle will be helpful for recognition. 
  The multi-scale training accuracy for combined spatial information and texture details (DEP) is 82.18\%. That's 9.02\% better than only focusing on spatial information (ResNet) and 6\% better than only focusing on texture details (Deep-TEN). 
  To gain insight into why DEP outperforms ResNet and Deep-TEN for material recognition, we visualize the features before classification layers of ResNet, Deep-TEN and DEP with Barnes-Hut t-SNE \cite{van2014accelerating} . We randomly choose 10000 images from training set for the experiment. The result is shown in Figure \ref{fig:tsne}. Notice that DEP separates classes farther apart and each class is clustered more compactly.

  \paragraph{Evaluation on MINC and DTD Dataset} To show the generality of DEP for material recognition, we experiment on two other material/texture recognition datasets: Describable Textures Database (DTD) \cite{cimpoi2014describing} and Materials in Context Database (MINC) \cite{bell2015material}. For an equal comparison, we build DEP based on a 50-layer ResNet \cite{he2016deep}, the feature maps channels from CNN streams are reduced from 2048 to 512 with a 1$\times$1 convolutional layer. The result is shown in Table~\ref{table:state-art}, DEP outperforms the state-of-the-art on both datasets. 
  Note that we only experiment with single scale training. As mentioned in \cite{lin2015bilinear}, multi-scale training is likely to improve results for all methods.

\section{Texture Manifold}
  
\begin{figure}[t]
\centering
\includegraphics[width= \linewidth]{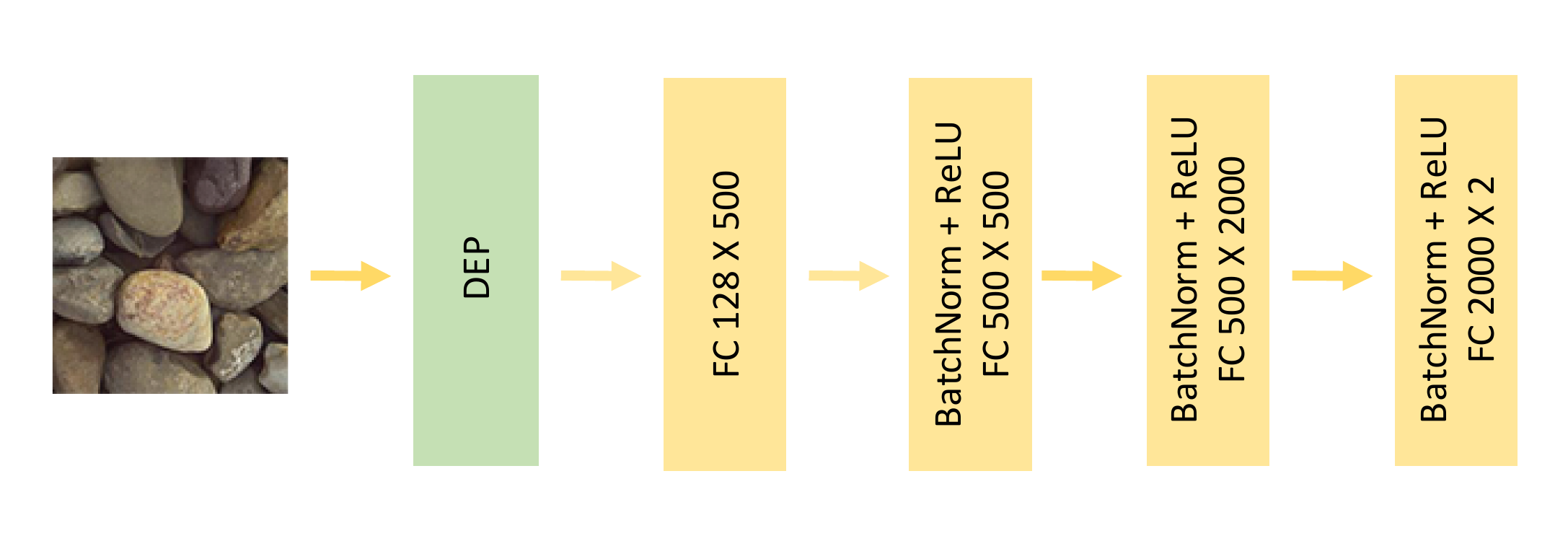}
\caption{The deep network for texture manifold, we employ DEP as feature extractor, outputs from the last fully connected layer of DEP works as input for texture embedding.}
\label{fig:tsne_model}
%\vspace{-0.2in}
\end{figure}

\begin{figure*}[t]
\centering
%\subfloat[Barnes-Hut t-SNE]
%{
%\includegraphics[width=.23\linewidth]{figure/bh_gtos_train.png}
%}
%\subfloat[Parametric t-SNE]
%{
%\includegraphics[width=.3\linewidth]{figure/img_param.png}
%}
\subfloat[DEP-parametric t-SNE]
{
\includegraphics[width=.45\linewidth]{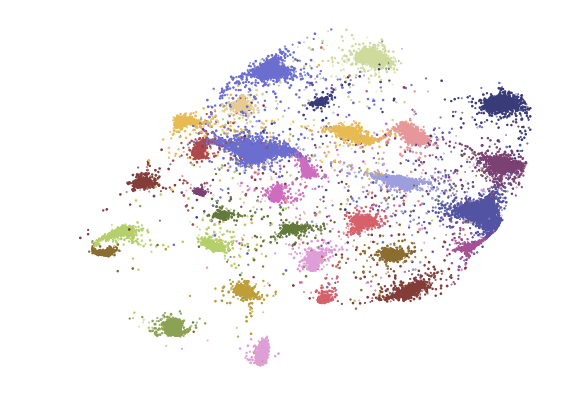}
}
\subfloat[DEP-manifold]
{
\includegraphics[width=.45\linewidth]{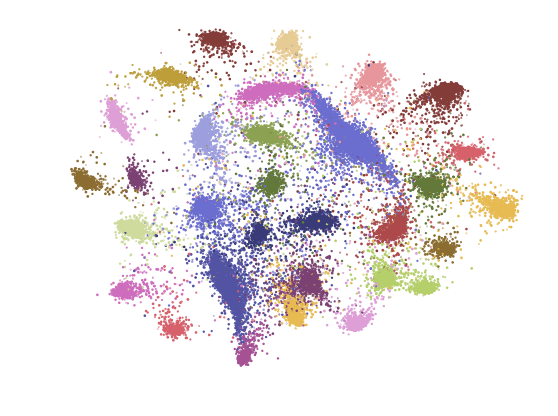}
}
\caption{Comparison the performance between DEP-parametric t-SNE and DEP-manifold with 60000 images from multi-scale GTOS dataset. For the embedded distribution of DEP-Parametric
t-SNE, the classes are distributed unevenly with crowding in some areas and sparseness in others. The DEP-manifold
has a better distribution of classes within the 2D embedding.}
\label{fig:tsne_predict}
%\vspace{-0.2in}
\end{figure*}

  Inspired by Parametric t-SNE \cite{van2009learning} and supervised t-SNE \cite{min2010deep,min2017exemplar}, we introduce a parametric texture manifold approach that learns to approximate the embedded distribution of non-parametric embedding algorithms \cite{maaten2008visualizing, van2014accelerating} %as input to train 
  using a deep neural network to directly predict the 2D manifold coordinates for the texture images.
  %We named our method DEP-manifold.
  We refer to this manifold learning method using DEP feature embedding as DEP-manifold. %, as the features used in the embedding are from based on the DEP pipeline. 
  Following prior work \cite{min2010deep,van2009learning}, the deep neural network structure is depicted in Figure ~\ref{fig:tsne_model}. Input features are the feature maps before the classification layer of DEP, which means each image is represented by a 128 dimensional vector. Unlike the experiment in \cite{min2010deep,van2009learning}, we add non-linear functions (Batch Normalization and ReLU) before fully connected layers, and we do not pre-train the network with a stack of Restricted Boltzmann Machines (RBMs) \cite{hinton2006fast}. We train the embedding network from scratch instead of the three-stage training procedure (pre-training, construction and fine-tuning) in parametric t-SNE and supervised t-SNE. We randomly choose 60000 images from the multi-scale GTOS dataset for the experiment. We experiment with DEP-parametric t-SNE, and DEP-manifold based on outputs from the last fully connected layer of DEP.

  \paragraph{Implementation} For the DEP-manifold, we employ Barnes-Hut t-SNE \cite{van2014accelerating} as a non-parametric embedding to build the embedded distribution. Following prior setting \cite{van2014accelerating}, we set perplexity to 30 and the output dimension of PCA to 50 for the experiment. For training the deep embedding network, we experiment with batch size 2048 and the parameters of the fully connected layers are initialized with the Xavier distribution~\cite{glorot2010understanding}. We employ L2 loss as the objective function for the experiment. The initial learning rate is 0.01, and decays by a factor of 0.1 every 30 epochs. The experiment is finished after 80 epochs. On an NVIDIA Titan X card, the training takes less than 5 minutes. 
  \paragraph{Texture Manifold}The texture manifold results are shown in Figure~\ref{fig:tsne_predict}. 
  %From the result we can see that parametric t-SNE doesn't work for material images.
  For the embedded distribution of DEP-Parametric t-SNE, the classes are  distributed unevenly with crowding in some areas and  sparseness in others.  %in the center and far away to the far-off points. So DEP-Parametric t-SNE has the crowding problem. 
 The  DEP-manifold has a better distribution of classes within the 2D embedding. We illustrate the texture manifold embedding by  randomly choosing 2000 images from training set to get the embedded distribution; then we embed images from test set into the DEP-manifold. Note that the test set images are not used in the computation of the DEP-manifold. The result is shown in Figure~\ref{fig:manifold}.  By observing the texture manifold, we find that for some classes, although the recognition accuracy is not perfect, the projected image is within the margin of the correct class, such as cement and stone-cement. Based on the similarity of material classes on the texture manifold, we build the confusion matrix for material recognition algorithms as shown in Figure~\ref{fig:tsne}.  For visualization, the one dimensional ordering of the confusion matrix axes are obtained from a one-dimensional embedding of the 2D manifold so that neighboring classes are close. Observe that for the DEP recognition (Figure~\ref{fig:tsne} c), there are very few off-diagonal elements in the confusion matrix. And the off-diagonal elements are often near diagonal  indicating  find when these images are misclassified,  they are recognized as closely-related classes.

%\begin{figure*}[t]
%\centering
%\subfloat[ResNet]
%{
%\includegraphics[width=.3\linewidth]{figure/resnet_conf.png}
%}
%\subfloat[Deep-TEN]
%{
%\includegraphics[width=.3\linewidth]{figure/deepten_conf.png}
%}
%\subfloat[DEP ({\small\TextRed{ours}})]
%{
%\includegraphics[width=.3\linewidth]{figure/bilinear_conf.png}
%}
%\caption{The confusion matrix of  ResNet (left), Deep-TEN (mid) and DEP (right). Classes are reordered based on the class similarity of Figure~\ref{fig:manifold}. We can find that some classes is misclassified, however, they are recognized as their related classes.}
%\label{fig:conf}
%\end{figure*}

\section{Conclusion}

We have developed methods for  recognition of ground terrain for potential applications in robotics and  automated vehicles.
We make three significant contributions in this paper: 1) introduction of Deep Encoding Pooling network (DEP) that leverages an orderless representation and local spatial information for recognition; 2) Introduction of DEP-manifold that integrates DEP network on top of a deep neural network to predict the manifold coordinates of a texture directly; 3) Collection of the GTOS-mobile database comprised of 81 ground terrains videos of similar terrain classes as GTOS, captured with a handheld mobile phone to evaluate knowledge-transfer between different image capture methods but within the the same domain.

\section*{Acknowledgment}
This work was supported by National Science Foundation 
award IIS-1421134. A TITAN X used for this research was 
donated by the NVIDIA Corporation.

{\small
\bibliographystyle{ieee}
\bibliography{egbib}
}

\end{document}